\documentclass[aoas]{imsart}

\RequirePackage{amsthm,amsmath,amsfonts,amssymb}
\RequirePackage[authoryear]{natbib}
\RequirePackage[colorlinks,citecolor=blue,urlcolor=blue]{hyperref}
\RequirePackage{graphicx}

\usepackage{algorithm}
\usepackage{algpseudocode}
\usepackage{paralist}
\usepackage{listings}
\usepackage{xcolor}
\definecolor{bg}{rgb}{0.95,0.95,0.92}

\startlocaldefs
\theoremstyle{plain}

\theoremstyle{definition}

\endlocaldefs

\makeatletter
\def\journal@name{}
\def\info@line{}
\makeatother

\begin{document}

\begin{frontmatter}
\title{Causal Diffusion Models for Counterfactual Outcome Distributions in Longitudinal Data}
\runtitle{Causal Diffusion Models}

\begin{aug}
\author[A]{\fnms{Farbod}~\snm{Alinezhad} \ead[label=e1]{alinezhad.f@northeastern.edu}},
\author[B]{\fnms{Jianfei}~\snm{Cao}\ead[label=e2]{j.cao@northeastern.edu}},
\author[C]{\fnms{Gary J.}~\snm{Young}\ead[label=e3]{ga.young@northeastern.edu}},
\and 
\author[D]{\fnms{Brady}~\snm{Post}\ead[label=e4]{b.post@northeastern.edu}}
\address[A]{Analysis Group\printead[presep={ ,\ }]{e1}}

\address[B]{Department of Economics,
Northeastern University\printead[presep={,\ }]{e2}}

\address[C]{Center for Health Policy and Healthcare
	Research, Northeastern University\printead[presep={,\ }]{e3}}

\address[D]{Department of Public Health and Health Sciences, Bouvé College of Health Sciences, Northeastern University\printead[presep={,\ }]{e4}}

\end{aug}

\begin{abstract}
Predicting counterfactual outcomes in longitudinal data, where sequential treatment decisions heavily depend on evolving patient states, is critical yet notoriously challenging due to complex time-dependent confounding and inadequate uncertainty quantification in existing methods. We introduce the Causal Diffusion Model (CDM), the first denoising diffusion probabilistic approach explicitly designed to generate full probabilistic distributions of counterfactual outcomes under sequential interventions. CDM employs a novel residual denoising architecture with relational self-attention, capturing intricate temporal dependencies and multimodal outcome trajectories without requiring explicit adjustments (e.g., inverse-probability weighting or adversarial balancing) for confounding. In rigorous evaluation on a pharmacokinetic–pharmacodynamic tumor-growth simulator widely adopted in prior work, CDM consistently outperforms state-of-the-art longitudinal causal inference methods, achieving a 15–30\% relative improvement in distributional accuracy (1-Wasserstein distance) while maintaining competitive or superior point-estimate accuracy (RMSE) under high-confounding regimes. By unifying uncertainty quantification and robust counterfactual prediction in complex, sequentially confounded settings, without tailored deconfounding, CDM offers a flexible, high-impact tool for decision support in medicine, policy evaluation, and other longitudinal domains.
\end{abstract}


\end{frontmatter}

\section{Introduction}

Counterfactual outcome prediction in longitudinal settings is a core problem in causal inference, with applications ranging from personalized medicine to policy
evaluation \citep{shalit2017}. Given a patient’s history and a sequence of treatments, we aim to
predict counterfactual outcomes, i.e., what would happen under alternative
treatment strategies over time. This approach to causal inference is grounded in the Rubin Causal Model (RCM), which defines causal effects through potential outcomes. In this framework, each individual has a set of potential outcomes corresponding to different treatment scenarios, and the causal effect is conceptualized as the difference between these outcomes. However, since only one of these potential outcomes, the factual outcome, is observable for each individual, namely the one corresponding to the treatment actually received, the other outcomes (counterfactual outcomes) remain unobserved, leading to the “fundamental problem of causal inference” \citep{rubin2005}.

While there are methods for predicting counterfactual outcomes in longitudinal settings, these methods are often designed to produce point
estimates of these outcomes (e.g., an expected value), but such point
predictions fail to capture uncertainty and the full range of plausible
trajectories \citep{bica2019,lim2018,melnychuk2022}. 
This is problematic in high-stakes domains like healthcare: if a
model does not quantify uncertainty in its counterfactual estimates, its utility
for decision-making is severely limited. For instance, clinicians not only need
an outcome estimate but also a sense of confidence or risk (e.g. probability of
adverse events) associated with a treatment decision 
\citep{begoli2019}. 
Furthermore, the presence of time-dependent confounding (where treatment assignment depends on past outcomes) complicates counterfactual prediction in this setting, as it introduces bias that must be accounted for in the model \citep{mortimer2005}.

To address this gap, we propose a diffusion-based generative model for
counterfactual prediction in longitudinal data. 
In this paper, we introduce Causal Diffusion Models (CDM) for counterfactual
time-series prediction. We design a diffusion architecture that iteratively
refines noisy outcome trajectories into realistic forecasts conditioned on
patient history and planned interventions. 
Diffusion models are a class of
probabilistic generative models that learn to sample complex distributions by
iterative denoising of random noise \citep{ho2020}. They have achieved state-of-the-art results
in modeling high-dimensional distributions (for images, audio, etc.) and have
recently been applied to time-series imputation and forecasting \citep{tashiro2021,alcaraz2023}.
Through this approach, we can
generate counterfactual outcome distributions that account for uncertainty and
temporal dependencies, directly addressing the limitations of point-estimate
models. 

Our contributions are threefolds.
First, we present the first
	use of diffusion probabilistic models for causal inference in longitudinal data,
	enabling the modeling of full conditional distributions of counterfactual outcomes rather than
	point estimates. This provides richer information (e.g. prediction intervals)
	and addresses the noted lack of uncertainty quantification in prior
	individualized treatment effect models.
	
	Second, we propose a novel
	diffusion model architecture with residual layers and skip connections tailored
	for time-series counterfactual prediction. To capture complex temporal
	interactions, we incorporate Relational Self-Attention (RSA) mechanisms inspired
	by video understanding models, allowing us to leverage relationships
	among time-varying covariates and treatments when denoising \citep{kim2021}.
	
Third, we perform extensive experiments on simulated longitudinal datasets with varying degrees of time-dependent confounding. Under low‐confounding regimes, CDM matches or slightly trails the best point‐estimate methods in RMSE, but as confounding strength increases, it consistently outperforms all baselines. Crucially, in distribution‐level evaluations using 1-Wasserstein distance, our model’s full predictive distributions are substantially closer to the ground truth than those obtained by applying Monte Carlo dropout to point‐estimate methods, underscoring the benefit of modeling the complete counterfactual outcome distribution.

Collectively, these advances establish CDM as the first method to deliver both state‑of‑the‑art point accuracy and highly accurate distributional forecasts in longitudinal causal inference, decisively pulling ahead of existing deep‑learning baselines. Our work combines advances in generative modeling with causal inference to improve counterfactual predictions over time. 

\medskip

\noindent\emph{Related Work.}
Our work contributes to the literature on counterfactual prediction in longitudinal data. Prior works have
tackled the challenge of estimating treatment effects over time in the presence
of time-dependent confounders using deep learning methods. \citet{lim2018} introduced the Recurrent Marginal
Structural Network (R-MSN), which uses a sequence-to-sequence LSTM architecture
along with marginal structural model theory (propensity score weighting) to
adjust for time-varying confounders. By reweighting training sequences to
simulate a randomized trial, R-MSN learns unbiased estimates of a patient’s
expected response to a sequence of treatments and showed improved accuracy on
simulated medical data. Building on representation learning, \citet{bica2019}
proposed the Counterfactual Recurrent Network (CRN). CRN employs an
encoder-decoder RNN coupled with domain adversarial training to learn balanced
representations of patient history. At each time step, it produces a hidden
state that is invariant to treatment assignment, breaking the association
between past health status and treatment choice. This results in a
representation that can be used for counterfactual outcome prediction as if
treatments were assigned randomly. CRN demonstrated lower error in counterfactual
estimation under time-varying confounding compared to earlier methods. Similarly, \citet{li2020} developed GNet, a deep learning implementation of g-computation, which uses recurrent neural networks to estimate the joint distribution of covariates and outcomes, allowing simulation of counterfactual trajectories under specified treatment plans. This approach explicitly models the evolution of outcomes under varying treatment scenarios, providing robust counterfactual predictions despite complex longitudinal confounding.
For medical applications relevant to our paper, see also \cite{ghali2023uq}, \cite{kang2021sleep}, \cite{logan2019bart}, \cite{riley2025risk}, \cite{schwarz2024uncertainty}, and \cite{tsaneva2025decoding}.

More recently, attention-based architectures have been explored for this
problem. \citet{melnychuk2022} introduced the Causal Transformer (CT) to capture
complex long-range dependencies in patient trajectories. Their model uses
separate transformer subnetworks for time-varying covariates, treatments, and
outcomes, which interact via cross-attention, and a custom training loss to
handle confounding. In particular, they propose a counterfactual domain
confusion loss that adversarially encourages the model’s latent representation
to be predictive of future outcomes while being non-predictive of the current
treatment, ensuring balance. This transformer-based approach achieved
state-of-the-art performance in comparison to the above methods,
highlighting the benefit of attention mechanisms for modeling temporal treatment
effects.
See also \cite{shi2025terra}, \cite{wang2024mamba}, \cite{wu2024counterfactual}, and \cite{xiong2024gtransformer}.

In addition, our work is related to the literature on diffusion models for time-series. 
Diffusion probabilistic models (also
known as score-based generative models) have rapidly gained popularity for
modeling complex data distributions \citep{chen2023imdiffusion,kollovieh2023prs,li2024d3u,rasul2021timegrad,zhou2024redi}. 
For a review, see \cite{su2025survey}.
In the time-series domain, \citet{tashiro2021}
introduced CSDI (Conditional Score-based Diffusion for Imputation), which
applies diffusion models to fill in missing values in time-series data. CSDI
conditions the denoising process on observed portions of the sequence,
explicitly training the model to exploit correlations and yield realistic
imputations. Notably, it outperformed conventional deterministic and variational
imputation methods by a large margin, demonstrating the power of diffusion
approaches for probabilistic time-series tasks.

Building upon this, \citet{alcaraz2023} proposed SSSD (Structured State Space Diffusion), which integrates diffusion models with structured state space architectures to handle long-term dependencies in time-series data. SSSD demonstrated state-of-the-art performance in both imputation and forecasting tasks across various datasets and missingness scenarios, including challenging blackout-missing situations where prior approaches struggled. Further extending the applicability of diffusion models, \citet{li2023} introduced Transformer-Modulated Diffusion Models (TMDM), combining transformer architectures with conditional diffusion generative processes for probabilistic multivariate time-series forecasting. By effectively leveraging transformers to capture complex temporal dependencies, TMDM showed superior distributional forecasting accuracy compared to existing deep forecasting approaches.

Our work extends diffusion-based modeling to causal forecasting \citep[see also][]{karimi2024dcrl,xia2025catsg}.
Unlike methods such as CSDI that assume a stationary data-generating process, we explicitly condition on interventions to account for time-dependent confounding. To capture structured interactions among covariates, treatments, and outcomes, across time, we draw on Relational Self-Attention (RSA) \citep{kim2021}. RSA was originally developed for video understanding, where it learns dynamic relational kernels that model dependencies across frames; here, we integrate an RSA-style module into our denoising blocks to learn interactions across multiple time series in longitudinal data.

\medskip


Section \ref{sec problem} formalizes the longitudinal causal forecasting problem.
Section \ref{sec cdm} introduces our Causal Diffusion Model (CDM), including the diffusion architecture and relational self-attention design. 
Section \ref{sec experiment} describes the PK–PD tumor-growth simulation environment and experimental setup, while Section \ref{sec result} presents the empirical results comparing CDM with state-of-the-art baselines. 
Section \ref{sec implementation} provides implementation details, and Section \ref{sec conclusion} concludes.

\section{Problem Formulation}
\label{sec problem}

We consider a longitudinal observational dataset with $N$ individuals (e.g., patients). For each individual $i$ and time step $t$ (where $t=1,2,\dots,T$), we observe: (1) a feature vector $X_{i,t}$ representing time-varying covariates (which may include prior outcomes or relevant clinical measurements), (2) a treatment $A_{i,t}$ administered at time $t$ (this could be a binary treatment, a dose, or a categorical choice among interventions), and (3) an outcome $Y_{i,t}$ of interest (e.g., a health metric). We denote the history of individual $i$ up to time $t_0$ by
\[
H_{i,0:t_0}
:=
\{(X_{i,t},A_{i,t},Y_{i,t})\}_{t=1}^{t_0}.
\]
In a causal inference framework, let $Y_{i,t}^{(a_{1:t})}$ be the potential outcome at time $t$ under the treatment sequence $a_{1:t}=(a_1,\dots,a_t)$.  Our goal is to estimate the counterfactual trajectory $\{Y_{i,1:T}^{(a_{1:T})}\}$ for a planned treatment sequence $a_{1:T}$, given past history $H_{i,0:t_0}$ (where $t_0=0$ corresponds to baseline).

A major challenge is time-varying confounding.  Unlike in a randomized trial, here $A_{i,t}$ can depend on past covariates or outcomes, which themselves affect future outcomes.  For instance, clinicians may prescribe a drug only when biomarkers exceed a threshold, and those biomarkers also influence patient risk.  Ignoring this can bias any naive predictor.

Formally, 
we aim to learn the full conditional distribution
\[
P\bigl(Y_{i,t_0+1:T}^{(a_{t_0+1:T})}\;\big|\;H_{i,0:t_0}\bigr),
\]
rather than merely its expectation
\(\mathbb{E}[Y_{i,t}^{(a_{1:t})}\mid H_{i,0:t_0}].\)  Modeling the entire distribution lets us answer queries such as “What is the probability that the patient’s blood pressure stays in a safe range under plan X?”  Capturing this richness is challenging because of the high dimensionality of multi-step outcomes and the stochasticity in patient responses.

Throughout, we will assume sequential ignorability. 

\section{A Causal Diffusion Model with Relational Self-Attention}
\label{sec cdm}

We now detail our proposed approach, Causal Diffusion Models (CDM), for
counterfactual outcome distribution prediction. The core idea is to use a
denoising diffusion probabilistic model to generate future outcomes under a
specified treatment plan, conditioned on the past history. Instead of directly
predicting an outcome value, our model learns a mapping from a known
distribution (e.g., a standard Normal distribution) to the distribution of
plausible future outcomes, capturing the uncertainty and variability in patient
responses to the specified treatments.

\subsection{Diffusion model overview} Diffusion models define a forward process
that gradually adds noise to data and a learnable reverse process that removes
the noise step by step. Here, we let $Y^{(a)}_{t_0+1:T}$ be the true future
outcome trajectory (given a planned treatment sequence $a$) from $t_0+1$ to $T$.
The forward process generates a sequence of noisy variables
$\{\mathbf{z}_1, \mathbf{z}_2, \dots, \mathbf{z}_K\}$, with each step adding a
small Gaussian perturbation. Formally,
\[
\mathbf{z}_k \;=\;\sqrt{1-\beta_k}\,\mathbf{z}_{k-1} \;+\;\sqrt{\beta_k}\,\boldsymbol{\varepsilon}_k,
\]
where $\beta_k$ is a small variance term, $\boldsymbol{\varepsilon}_k$ is
standard Gaussian noise, and $\mathbf{z}_0 = Y^{(a)}_{t_0+1:T}$. The parameter
sequence $\{\beta_k\}$ (sometimes called the “noise schedule”) determines how
rapidly information is noised. In practice, we adopt a cosine schedule for
$\{\beta_k\}$ owing to its favorable empirical performance with fewer diffusion
steps \citep{nichol2021}.

Reversing this noising process involves a denoising function
$\mathbf{f}_\theta(\mathbf{z}_k,k)$ that predicts the added noise at step $k$.
At test time, we start from $\mathbf{z}_K\sim \mathcal{N}(0,I)$ and repeatedly
apply $\mathbf{f}_\theta$ to remove noise, culminating in $\mathbf{z}_0$ as a
sample from the (conditional) distribution of interest. The conditioning here
includes the patient’s observed history $H_{i,t_0}$ up to $t_0$ and the planned
treatment sequence $a_{t_0+1:T}$: we incorporate these via context embeddings
and a mask that fixes any known observations (so only truly missing or
counterfactual dimensions are exposed to the diffusion process).

\subsection{The proposed algorithm} 

The model $\mathbf{f}_\theta(\cdot)$
takes as input: (a)~the current noisy signal $\mathbf{z}_k$ (shaped as
$\text{batch}\times\text{time}\times\text{features}$), (b)~the diffusion step
index $k$ (encoded via a learnable embedding), and (c)~a mask indicating which
coordinates remain unknown and thus receive noise. Internally, we further embed
time indices (so the network can differentiate earlier vs.\ later time steps)
and feature indices (for each dimension of the outcome and history features). The
network processes these embeddings via a stack of residual blocks that apply
Relational Self-Attention (RSA) across the time and feature axes.

To learn the parameters $\theta$, we draw a sample $Y^{(a)}_{t_0+1:T}$ from our
training set, pick a random diffusion step $k$, and inject noise into the
unobserved coordinates to obtain $\mathbf{z}_k$. The model predicts the noise,
$\mathbf{f}_\theta(\mathbf{z}_k,k)$, and we compare this to the true noise
$\boldsymbol{\varepsilon}$ that generated $\mathbf{z}_k$. We minimize the
mean-squared error between these quantities, restricted to masked (i.e. missing
or counterfactual) entries:
\[
\mathcal{L}(\theta) \;=\;\text{E}_{k,\mathbf{z}_k}\,(\|\mathbf{f}_\theta(\mathbf{z}_k,k)\;-\;\boldsymbol{\varepsilon}_k)\|^2.
\]
Training thus teaches $\mathbf{f}_\theta$ to remove noise specifically where the
data is unknown and maintain consistency with observed entries.


\begin{algorithm}[h!]
	\caption{RSA-Based Diffusion Training}
	\label{alg:training_diffusion}
	\begin{algorithmic}[1]
		\Require Training data $\{x_i\}$, mask strategy, diffusion steps $K$, RSA-based model $f_\theta$
		\Ensure Learned parameters \(\theta\)
		\State Initialize \(\theta\)
		\For{each epoch}
		\ForAll{each batch \(x\)}
		\State \(M \leftarrow \text{get\_mask}(x)\)
		\State Sample diffusion step \(k\) and noise \(\epsilon \sim \mathcal{N}(0,I)\)
		\State Compute noised data \(x_k \leftarrow \sqrt{\alpha_k}\,x + \sqrt{1-\alpha_k}(\epsilon \odot M)\)
		\State Predict noise: \(\hat{\epsilon}_\theta \leftarrow f_\theta(x_k, M, k)\)
		\State Compute loss: \(\mathcal{L} \leftarrow \|(\hat{\epsilon}_\theta - \epsilon)\odot M\|^2\)
		\State Update parameters: \(\theta \leftarrow \theta - \nabla_\theta\mathcal{L}\)
		\EndFor
		\EndFor \\
		\Return Learned parameters \(\theta\)
	\end{algorithmic}
\end{algorithm}

\subsection{Sampling and prediction:} After training, the model can generate a
counterfactual trajectory by:
\begin{enumerate}
	\item Initializing $\mathbf{z}_K$ with random samples from a standard Gaussian distribution.
	\item Iterating the reverse diffusion process from $k=K$ down to $k=1$, using
	the learned function $\mathbf{f}_\theta(\mathbf{z}_k, t_k, C)$ to iteratively
	denoise and sample.
\end{enumerate}
Because each run of the reverse diffusion is stochastic, repeating the process
yields a sample from the distribution of plausible counterfactual trajectories.

\subsection{Assumptions}

Throughout, we assume sequential ignorability: that the potential outcomes are independent of the treatment assignment given the observed history. 
This is standard as in the baseline causal inference methods (e.g., R-MSN, CRN, CT). 
We also assume consistency: for any individual $i$ and time $t$, the observed outcome equals the potential outcome under the treatment history actually received. Finally, we assume positivity: that each treatment has a non-zero probability of being assigned to each individual.

\section{Experiment Design}
\label{sec experiment}

Since in real-world settings we lack ground-truth counterfactual outcomes, we evaluate our model on simulated data to have access to the true counterfactual outcomes.
Similar to the other studies on this topic, we use the pharmacokinetic/pharmacodynamic (PK-PD) tumor growth simulator \citep{geng2017}. This state-of-the-art simulator generates realistic longitudinal patient data with evolving tumor sizes and treatments and provides us with “ground-truth” counterfactual outcomes, allowing quantitative
evaluation of both factual and counterfactual predictions under varying degrees
of time-dependent confounding. Below, we describe the dataset, the simulation
procedure in detail (including key equations), and the evaluation metrics.

\subsection{PK-PD tumor growth simulation}
We simulate a cohort of virtual patients, each with evolving tumor size. At each
time step~\(t\), a patient can receive one of several treatments (chemotherapy,
radiotherapy, or both), which in turn affects tumor growth/shrinkage at the next
time step. Crucially, the treatment assignment is \emph{confounded} by tumor
progression: for instance, if the tumor is growing quickly, there is a higher
probability that the patient will receive more aggressive therapy. This
introduces time-varying confounding, which we control via a parameter~\(\gamma\).
Larger~\(\gamma\) indicates a steeper logistic relationship between tumor size
and treatment assignment probability, thus exacerbating the confounding.

\subsubsection{Initial Tumor Sizes and Patient Heterogeneity}
Each patient begins with an initial tumor size sampled from a truncated
log-normal distribution, stratified by cancer stage (I, II, III, or IV). We
assume the tumor is roughly spherical, so we track its \emph{volume}~\(V_t\).
Patient-level parameters \(\{\alpha, \rho, \beta, \beta_c\}\) are drawn from
correlated distributions to capture different underlying growth and
treatment-response rates:
\begin{itemize}
	\item \(\rho\) governs the baseline tumor growth,
	\item \(\alpha,\beta\) govern radiotherapy response,
	\item \(\beta_c\) governs the chemotherapy effect,
	\item \(K\) is a carrying capacity parameter (fixed across patients).
\end{itemize}
We also include random noise~\(\epsilon_t\) at each step for
inter-individual variability.

\subsubsection{Tumor Growth Equation}
Between time steps, if \(V_t\) is the tumor volume at time~\(t\), the next
volume~\(V_{t+1}\) is updated according to:
\[
V_{t+1} \;=\; V_t \;([\,
1 \;+\;\rho\,\log\!((\tfrac{K}{V_t}))
\;-\;\beta_c \,\text{Chemo}_t
\;-\;\bigl(\alpha\,\text{Radio}_t \;+\;\beta\,\text{Radio}_t^2\bigr)
\;+\;\epsilon_t)],
\]
where \(\text{Chemo}_t\) and \(\text{Radio}_t\) represent effective dosage
levels (accounting for accumulation or decay) of chemotherapy and radiotherapy
at time~\(t\). We cap~\(V_{t+1}\) at a maximum threshold (the “death threshold”)
and allow for the possibility of spontaneous “recovery” (i.e., the tumor volume
dropping to zero with a small probability).

\subsubsection{Time-Dependent Confounding via Logistic Assignment}
To simulate realistic treatment policies, we specify that at each step,
clinicians choose whether to administer chemotherapy and/or radiotherapy based
on the \emph{recent average tumor diameter} (over a sliding window).
Specifically, the probability of administering chemotherapy is:
\[
p(\text{Chemo}_t = 1) \;=\;
\sigma\!\bigl(\gamma_{\text{chemo}}\,[\,\text{Diameter}_\text{recent}\,-\,\text{Intercept}\,]\bigr),
\]
where \(\sigma\) is the sigmoid function. A similar logistic form with
\(\gamma_{\text{radio}}\) is used for radiotherapy. Larger values of~\(\gamma\)
make the model \emph{more sensitive} to tumor size, thus increasing the
magnitude of the time-varying confounding (i.e., sicker patients are more likely
to get treated). Within our experiments, we use the same \(\gamma\) for both chemo and radiotherapy.

\subsection{The proposed method}

After training our model on “factual” patient trajectories (similar to what would happen in a real-life scenario), we generate counterfactual outcomes by \emph{simulating} the next time step for our test patients under all possible treatment
choices (e.g., “no treatment,” “chemo only,” “radio only,” or “both”) at each time-step. Because
we have a stochastic model~\((\epsilon_t)\), each patient’s one-step-ahead
counterfactual is itself a distribution. We sample from this distribution to obtain multiple plausible counterfactual tumor
volumes for each treatment choice. This allows us to evaluate how well different
methods predict not just a single point estimate but the entire \emph{distribution}
of counterfactual outcomes.

For our CDM, we perform hyperparameter tuning over learning rate and embedding sizes, and we conduct a dedicated ablation study to assess key design choices: diffusion steps (5 vs. 20), Beta schedule (cosine vs. linear), inclusion of residual layers, and backbone architecture (RSA vs. a simple neural network). These ablations evaluate how each component contributes to both RMSE and 1-Wasserstein performance. All models are trained with the Adam optimizer (initial LR = 1e-3 with 0.5× decay on validation loss plateau) for 25 epochs and batch size 200; full hyperparameter settings are in Appendix~\ref{sec:appendix_hparams}. At test time, we draw 20 samples per counterfactual. We found that 20 samples were sufficient to achieve stable performance across all models, as larger sample sizes did not yield significant improvements in the evaluation metrics. All experiments were conducted on a node equipped with four NVIDIA A100 GPUs, requiring approximately 24 GPU-hours in total.

Due to limited computational resources, we evaluated CDM's seed-to-seed variability at three representative confounding levels (\(\gamma = 0,5,10\)) using four random seeds. We observed low variability (standard deviations below 0.01\% for both RMSE\(_\text{median}\) and 1-Wasserstein), which is negligible compared to the performance gaps between methods. Consequently, we report single-seed results for the remaining experiments.


For the baseline methods, we use the hyperparameters listed in the Causal
Transformer study by \citet{melnychuk2022} for $\gamma$ levels 0 to 5.
For $\gamma$ levels above that, we
use a hyperparameter tuning method similar to that described in the Causal Transformer study. We train these models using their original loss functions
(R-MSN with IPTW reweighting, CRN with adversarial loss, CT with domain
confusion and their training schedules).

As our training dataset, we create a cohort of 10{,}000 patients with 60 time
steps each, similar to the method used in both the CRN and CT. We generate 1{,}000
validation and 1{,}000 test patients for evaluation. For the test set, we
simulate counterfactual outcomes under different treatment choices at each of the time-steps, leading to \( (T-1) \times 4 \) counterfactual outcomes for each patient. We create
100 samples for each counterfactual outcome to approximate the counterfactual
distribution for each treatment choice. We evaluate the models
on the test set using the RMSE and Wasserstein distance metrics described
earlier. Importantly, we use the same training, test, and validation datasets
for all models to ensure a fair comparison. We repeat this process for different
levels of time-dependent confounding ($\gamma$) to assess the models’
performance under varying degrees of time-dependent confounding.


\subsection{Baselines}

We compare our diffusion model against several
state-of-the-art methods for longitudinal causal inference, each designed to
handle time-varying confounding in sequential data:

\begin{enumerate}
	\item {R-MSN} -- Recurrent Marginal Structural Network \citep{lim2018};
	\item {CRN} -- Counterfactual Recurrent Network \citep{bica2019};
	\item {GNet} -- Deep learning g-computation model \citep{li2020};
	\item {CT} -- Causal Transformer \citep{melnychuk2022}.
\end{enumerate}

For all baseline methods, which natively produce point estimates, we leverage Monte Carlo dropout at test time to approximate predictive uncertainty as recommended in their original publications \citep{gal2016}.

\subsection{Evaluation metrics}

We evaluate the predicted distribution over outcomes relative to the ground-truth using two metrics:
{RMSE (from Median)} and {1-Wasserstein Distance}.

To make both metrics comparable, we divide by a normalization constant (the maximum tumor volume threshold \(V_{\max}\)) and then multiply by \(100\). Thus, all reported values can be interpreted as percentages of the maximum possible tumor size.

Formally, let \(\hat{V}_{i,t,s}\) denote a model’s sampled volume (sample~\(s\) for patient~\(i\) at time~\(t\)), and let \(V_{i,t}\) be the ground-truth volume. We define a binary mask \(m_{i,t}\) indicating the indices for which a counterfactual volume \(Y^{(a)}_{t_0+1}\) is predicted. The standardized metrics are as follows:

1) {RMSE} (from Median).
For each patient \(i\) and time \(t\), let
\[
\widetilde V_{i,t}
:= \mathrm{median}_{s}\bigl(\hat V_{i,t,s}\bigr)
\quad\text{and}\quad
\widetilde V^*_{i,t}
:= \mathrm{median}_{s}\bigl(V_{i,t,s}\bigr),
\]
where \(\hat V_{i,t,s}\) are the model’s sampled volumes and \(V_{i,t,s}\) are the simulator’s sampled “ground‐truth” volumes. We then define
\[
\mathrm{RMSE}_{\mathrm{median}}
=
\frac{100}{V_{\max}}
\sqrt{
	\frac{1}{\sum_{i,t}m_{i,t}}
	\sum_{i,t}m_{i,t}\,\bigl(\widetilde V_{i,t} - \widetilde V^*_{i,t}\bigr)^2
}.
\]
This measures point accuracy: lower RMSE indicates the predicted medians align more closely with ground-truth.

We also compute the RMSE for the 2.5th and 97.5th percentiles of the predicted distribution using a similar approach to capture the spread of the predicted distribution and how well it captures the ground-truth distribution at the tails.

2) {1-Wasserstein Distance}.
For each patient \(i\) and time \(t\), let
\[
\mathbf{Q}_{i,t}
:=
\bigl(Q_1,\dots,Q_K\bigr)
=
\mathrm{Quantiles}\bigl(\{\hat V_{i,t,s}\}_{s=1}^S\bigr)
,\quad
\mathbf{Q}^*_{i,t}
:=
\mathrm{Quantiles}\bigl(\{V_{i,t,s}\}_{s=1}^S\bigr),
\]
where \(\{\hat V_{i,t,s}\}\) are the model’s \(S\) samples and \(\{V_{i,t,s}\}\) are the simulator’s \(S\) ground‐truth samples, and \(K\) is the number of quantile bins (we use \(K=100\)). We then define the normalized 1-Wasserstein distance as
\[
W_1
=
\frac{100}{V_{\max}}
\;\frac{1}{\sum_{i,t}m_{i,t}}
\sum_{i,t} m_{i,t}\,\bigl\lVert \mathbf{Q}_{i,t} - \mathbf{Q}^*_{i,t}\bigr\rVert_1.
\]
A lower \(W_1\) indicates that the full predicted distribution more closely matches the true counterfactual distribution.

Together, these metrics measure both \emph{point accuracy} (via RMSE) and \emph{distributional fidelity} (via Wasserstein distance), providing a comprehensive evaluation of how well the model recovers the true causal (counterfactual) relationships in the PK-PD simulator.

\section{Experiment Results}
\label{sec result}

We compare our Causal Diffusion Model (CDM) with four baselines---R-MSN, CRN,
GNet, and Causal Transformer (CT)---over a range of confounding levels
\(\gamma\). 

\subsection{Main findings}

\begin{figure}[t!]
	\centering
	\includegraphics[scale=.7]{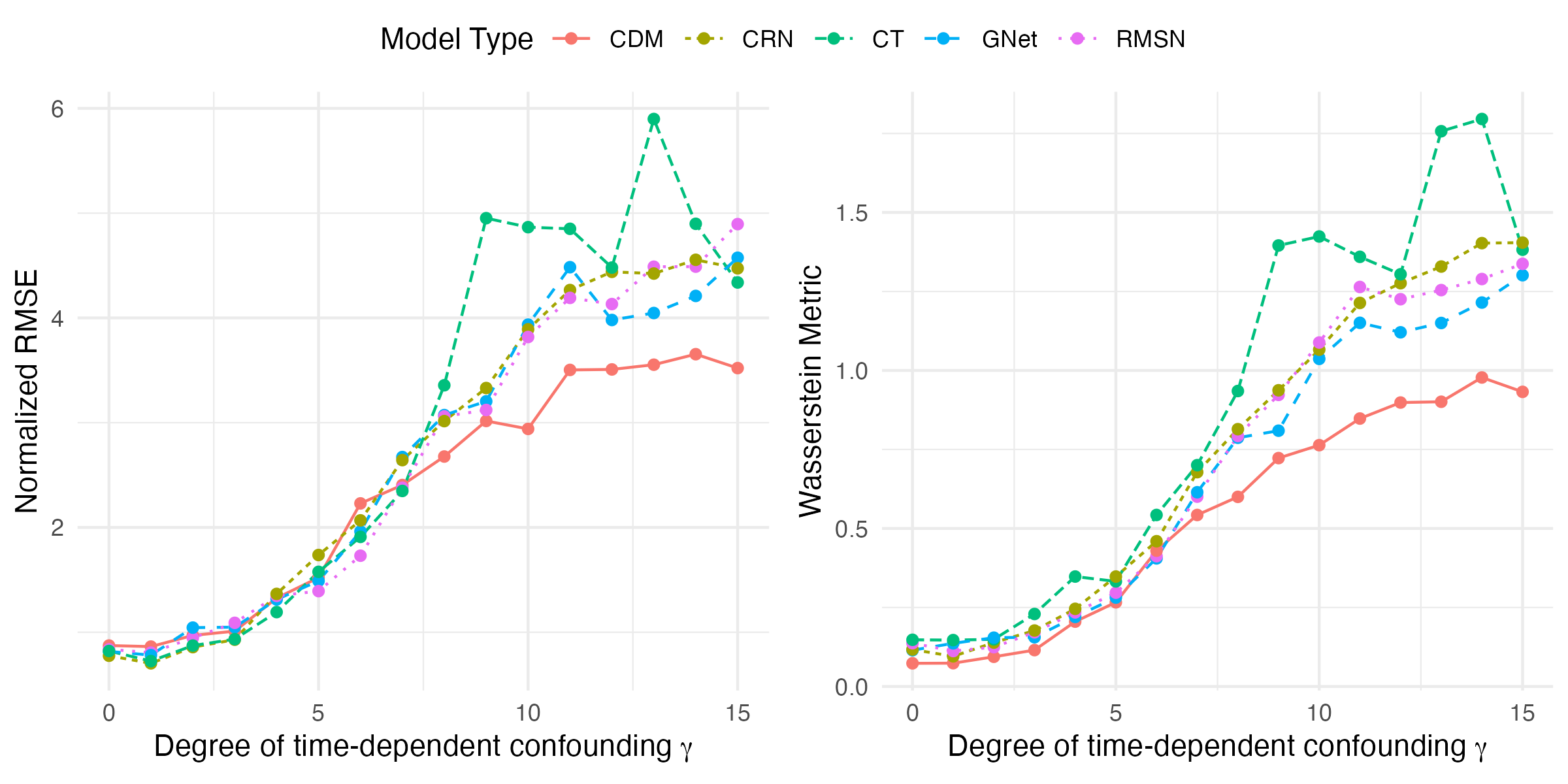}
	\caption{Root Mean Squared Error (RMSE, \%) and 1-Wasserstein distance (\%) vs. confounding level \(\gamma\).}
	\label{fig:rmse}
\end{figure}

Figure~\ref{fig:rmse} summarizes the key trends.
We observe that CDM is best {overall} in distribution fidelity.
A standout observation is that {CDM achieves the lowest Wasserstein
	distance \emph{for every}~\(\gamma\)}, from 0 (no confounding) up to 15 (heavy
confounding). Concretely, at minimal confounding (\(\gamma=0\)), CDM achieves a 1-Wasserstein of 0.08\% versus 0.11\% for the next-best method (GNet), a 27\% relative reduction. Under heavy confounding (\(\gamma=10\)–15), CDM’s distances range from 0.87\% to 0.91\%, compared to 1.04\%–1.30\% for the closest baseline, yielding 16–30\% relative improvements. Hence, \emph{across the entire confounding spectrum}, CDM consistently aligns more closely with the true counterfactual distributions.

In low and moderate confounding regimes (\(\gamma \leq 7\)), CDM achieves RMSE values competitive with the best baselines, often within \(0.1\%\) of the lowest error, demonstrating that its superior distributional fidelity does not sacrifice point‐estimate accuracy in these settings.

At high confounding levels (\(\gamma \geq 8\)), CDM ranks consistently best in
RMSE, sometimes by a substantial margin. Thus, if time-dependent confounding is severe, CDM may
offer the strongest overall performance in both point accuracy and distribution
estimation.

Overall, CDM consistently delivers the most faithful full distributions, achieving the lowest 1-Wasserstein distance across almost all confounding levels, while its RMSE is competitive with or superior to baselines, especially under heavy confounding where it overtakes them in point accuracy. Even when some methods slightly outperform CDM in RMSE at low-to-moderate confounding, these gains vanish or reverse as confounding intensifies.

\subsection{Detailed results}

\begin{table}[H]
	\centering
	\caption{Performance of CDM versus alternative methods (\%)}
	\label{tab:rmse}
	\setlength{\tabcolsep}{3.5pt}  
		\begin{tabular}{lcccccccccccccccc}
			\hline
			&			\multicolumn{16}{c}{Amount of Time-Dependent Confounding ($\gamma$)}                                                  \\
			\cline{2-17}
			& 0    & 1    & 2    & 3    & 4    & 5    & 6    & 7    & 8    & 9    & 10   & 11   & 12   & 13   & 14   & 15   \\
			\hline
			RMSE \\
			\hline
			CDM   & 0.88 & 0.89 & 0.95 & 1.61 & 1.50 & 1.78 & 2.14 & 2.41 & 2.69 & 2.95 & 3.62 & 3.43 & 3.45 & 4.09 & 3.50 & 3.73 \\
			CRN   & 0.78 & 0.70 & 0.86 & 0.93 & 1.37 & 1.74 & 2.07 & 2.64 & 3.01 & 3.33 & 3.89 & 4.27 & 4.44 & 4.42 & 4.55 & 4.47 \\
			CT    & 0.82 & 0.72 & 0.87 & 0.93 & 1.19 & 1.58 & 1.91 & 2.35 & 3.36 & 4.79 & 4.87 & 4.85 & 4.48 & 5.86 & 5.35 & 4.34 \\
			GNet  & 0.82 & 0.78 & 1.04 & 1.05 & 1.31 & 1.49 & 1.96 & 2.67 & 3.07 & 3.20 & 3.94 & 4.09 & 3.98 & 4.05 & 4.21 & 4.57 \\
			RMSN  & 0.84 & 0.80 & 0.95 & 1.09 & 1.35 & 1.39 & 1.73 & 2.38 & 3.06 & 3.12 & 3.82 & 4.19 & 4.13 & 4.49 & 4.49 & 4.90 \\
			\hline
			\multicolumn{2}{l}{Wasserstein} \\
			\hline
			CDM   & 0.08 & 0.07 & 0.08 & 0.14 & 0.20 & 0.26 & 0.37 & 0.50 & 0.61 & 0.75 & 0.87 & 0.92 & 0.83 & 0.98 & 0.95 & 0.91 \\
			CRN   & 0.12 & 0.10 & 0.14 & 0.18 & 0.25 & 0.35 & 0.46 & 0.68 & 0.81 & 0.94 & 1.07 & 1.21 & 1.28 & 1.33 & 1.40 & 1.40 \\
			CT    & 0.15 & 0.15 & 0.15 & 0.23 & 0.35 & 0.33 & 0.54 & 0.70 & 0.93 & 1.32 & 1.42 & 1.36 & 1.30 & 1.70 & 1.60 & 1.38 \\
			GNet  & 0.11 & 0.14 & 0.15 & 0.16 & 0.22 & 0.28 & 0.41 & 0.61 & 0.79 & 0.81 & 1.04 & 1.12 & 1.12 & 1.15 & 1.21 & 1.30 \\
			RMSN  & 0.14 & 0.11 & 0.12 & 0.17 & 0.23 & 0.30 & 0.41 & 0.60 & 0.79 & 0.92 & 1.09 & 1.26 & 1.23 & 1.25 & 1.29 & 1.34 \\
			\hline
		\end{tabular}
\end{table}
Tables~\ref{tab:rmse} shows the complete numeric results (RMSE and
1-Wasserstein distance, both given as percentages of the maximum possible tumor
size).
The tables show that CDM consistently
outperforms the baselines in terms of Wasserstein distance, capturing the
distribution of counterfactual outcomes more accurately. The RMSE results are
more nuanced, with CDM sometimes ranking lower than other methods at moderate
levels of confounding, but consistently outperforming them at higher levels of
confounding.

\subsection{Tail performance}

\begin{figure}[b!]
	\centering
	\includegraphics[scale=.7]{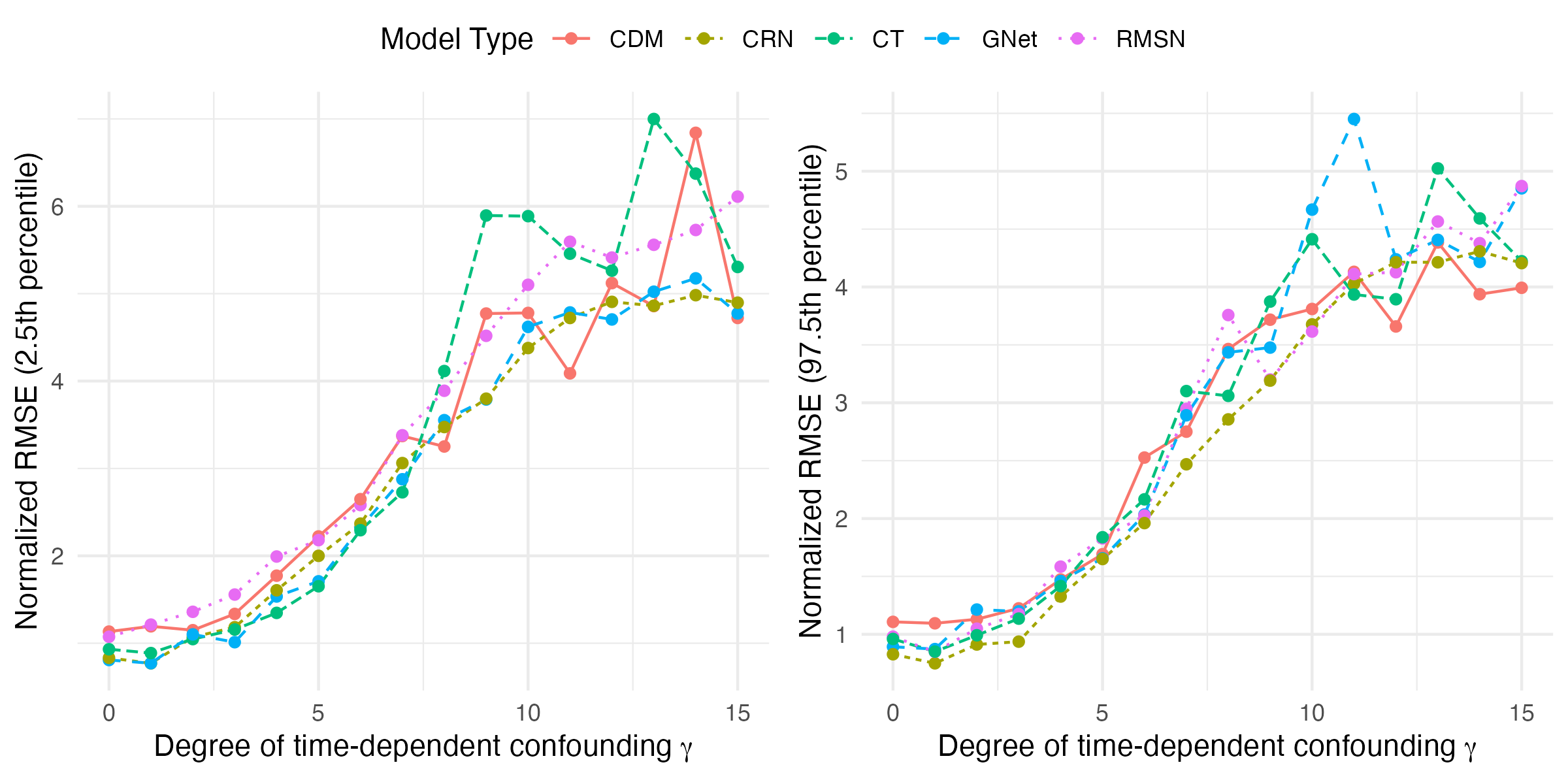}
	\caption{RMSE (\%) for the 2.5th and 97.5th percentiles of the predicted distribution. CDM (Causal Diffusion Model), CRN (Counterfactual Recurrent Network), CT (Causal Transformer), GNet, and RMSN (Recurrent Marginal Structural Network) are compared. CDM performs competitively with the baselines at low and moderate confounding levels, but outperforms most of the baselines at higher confounding levels.}
	\label{fig:rmse_tails}
\end{figure}

Figure~\ref{fig:rmse_tails} shows the RMSE for the 2.5th and 97.5th
percentiles of the predicted distribution, providing a more detailed view of the
model performance at the tails of the distribution.

\subsection{Ablation study}

\begin{table*}
	\caption{Ablation study on CDM design choices at three confounding levels.}
		\label{tab:ablation_full}
	\begin{tabular}{@{}lccccccc@{}}
		\hline
			& \multicolumn{3}{c}{\textbf{RMSE$_{\text{median}}$ (\%)}}&
					& \multicolumn{3}{c}{\textbf{1-Wasserstein (\%)}}                                                              \\
					\cline{2-4} \cline{6-8}
				Configuration                 & $\gamma{=}0$                                             & $\gamma{=}5$ & $\gamma{=}10$ &
				& $\gamma{=}0$                                             & $\gamma{=}5$ & $\gamma{=}10$                      \\
				\hline
				Full CDM (baseline)           & 0.94                                                     & 1.64         & 4.29       &   & 0.08 & 0.26 & 1.04 \\
				Diffusion steps $=20$         & 1.12                                                     & 3.37         & 3.24       &   & 0.09 & 0.68 & 0.83 \\
				Linear $\beta$ schedule       & 1.90                                                     & 4.53         & 8.53       &   & 0.13 & 0.65 & 1.91 \\
				Inclusion of a residual layer & 1.05                                                     & 1.87         & 3.19        &  & 0.09 & 0.33 & 0.83 \\
				Embedding dimension $=8$      & 2.90                                                     & 5.37         & 8.46       &   & 0.47 & 1.38 & 3.73 \\
				Simple‑NN backbone            & 3.78                                                     & 7.03         & 12.77     &    & 0.74 & 1.91 & 4.55 \\
				\hline
	\end{tabular}
\end{table*}

The results of our ablation study are reported in Table~\ref{tab:ablation_full}.  The results show that every component helps, with the cosine $\beta$ schedule and the RSA backbone providing the largest gains.

\section{Implementation Details}
\label{sec implementation}

\subsection{Python implementation}{\label{sec:appendix_implementation}}

\subsubsection{RSA multi-head attention: \texttt{TimesSeriesAttention}}
We implement RSA-based attention for processing longitudinal data shaped
\((B, T, F, d_{\text{in}})\), corresponding to batch size, time steps,
features, and input embedding dimension, respectively. Core operations involve
linear projection, relational kernel generation, and combination with
appearance-based features.

\begin{lstlisting}[language=Python, caption={RSA-based TimesSeriesAttention}]
	class TimesSeriesAttention(nn.Module):
	def __init__(self, d_in, d_out, nh=8, kernel_size=(3,7)):
	super().__init__()
	self.projection_linear = nn.Sequential(
	nn.Linear(d_in, d_out, bias=False),
	nn.SiLU(),
	nn.Linear(d_out, d_out, bias=False),
	)
	self.H1 = nn.Conv2d(...)
	self.H2 = nn.Conv2d(...)
	
	def forward(self, x):
	# Project input
	x_proj = self.projection_linear(x.permute(0,2,3,1))
	x_proj = x_proj.permute(0,3,1,2)
	
	# Split into Q, K, V and normalize
	q, k, v = torch.split(x_proj, [...], dim=1)
	q, k, v = self.L2norm(q), self.L2norm(k), self.L2norm(v)
	
	# Compute relational kernel
	kernel = torch.einsum("abc,abd->acd", q.transpose(1,2), self.H1(k))
	
	# Appearance-based feature extraction
	v_trans = self.H2(v)
	
	# Combine and reshape output
	feature = torch.einsum("abc,abd->acd", v_trans, kernel)
	out = feature.permute(0,2,3,1)
	return out
\end{lstlisting}

\subsubsection{Transformer encoder cells}
Each \texttt{TransformerEncoderCell} applies RSA attention followed by a
feed-forward layer, residual connections, and layer normalization.

\begin{lstlisting}[language=Python, caption={Transformer Encoder Cell}]
	class TransformerEncoderCell(nn.Module):
	def __init__(self, embed_dim, num_heads, kernel_size, ff_dim, dropout=0.1):
	super().__init__()
	self.attention = TimesSeriesAttention(embed_dim, embed_dim, num_heads, kernel_size)
	self.ff_layer = nn.Sequential(
	nn.Linear(embed_dim, ff_dim),
	nn.ReLU(),
	nn.Linear(ff_dim, embed_dim)
	)
	self.dropout = nn.Dropout(dropout)
	self.layer_norm = nn.LayerNorm(embed_dim)
	
	def forward(self, x):
	attn_out = self.attention(x)
	x = self.layer_norm(x + self.dropout(attn_out))
	ff_out = self.ff_layer(x)
	x = self.layer_norm(x + self.dropout(ff_out))
	return x
\end{lstlisting}

The full Transformer encoder stacks several such cells:

\begin{lstlisting}[language=Python, caption={Full Transformer Encoder}]
	class TransformerEncoder(nn.Module):
	def __init__(self, embed_dim, num_heads, kernel_size, ff_dim, num_cells):
	super().__init__()
	self.layers = nn.ModuleList([
	TransformerEncoderCell(embed_dim, num_heads, kernel_size, ff_dim)
	for _ in range(num_cells)
	])
	self.final_norm = nn.LayerNorm(embed_dim)
	
	def forward(self, x):
	for layer in self.layers:
	x = layer(x)
	return self.final_norm(x)
\end{lstlisting}

\subsubsection{Selective masking for counterfactual prediction}
We construct masks for selectively adding noise to specific data coordinates,
facilitating training with counterfactual or partially observed scenarios:
\begin{lstlisting}[language=Python, caption={Selective Masking Procedure}]
	def get_mask(data, features_to_impute, last_n_time, sequence_length):
	b, t, f = data.shape
	mask = torch.zeros_like(data)
	for i in range(b):
	seq_len = int(sequence_length[i])
	for j, feature in enumerate(features_to_impute):
	last_n = last_n_time[j]
	start_idx = max(seq_len - last_n, 0)
	mask[i, start_idx:seq_len, feature] = 1
	return mask
\end{lstlisting}

%
%

\subsection{Hyperparameter configuration}{\label{sec:appendix_hparams}}
This section provides a detailed list of hyperparameters used for training the
RSA-based diffusion imputer. Table~\ref{tab:train_hyperparams} outlines the
settings for the training procedure, while
Table~\ref{tab:diffusion_hyperparams} lists the configuration of the diffusion
imputation model.

\begin{table}[H]
	\centering
	\caption{Training Hyperparameters}
	\label{tab:train_hyperparams}
	\renewcommand{\arraystretch}{1.2}
	\begin{tabular}{l l}
		\hline
		\textbf{Hyperparameter} & \textbf{Value}                                                        \\
		\hline
		Gradient Clipping       & 4                                                                     \\
		Epochs                  & 25                                                                    \\
		Learning Rate (LR)      & starting at $10^{-3}$, decaying factor 0.9 on validation loss plateau \\
		Minimum Learning Rate   & $10^{-6}$                                                             \\
		\hline
	\end{tabular}
\end{table}

\begin{table}[H]
	\centering
	\caption{Diffusion Imputation Model Hyperparameters}
	\label{tab:diffusion_hyperparams}
	\renewcommand{\arraystretch}{1.2}
	\begin{tabular}{l l}
\hline
		\textbf{Hyperparameter}           & \textbf{Value}  \\
		\hline
		Embedding Dimension               & 32              \\
		Residual Layers                   & 2               \\
		Diffusion Steps                   & 5               \\
		Beta Schedule                     & \texttt{cosine} \\
		Attention Heads                   & 8               \\
		Kernel Size (for RSA convolution) & (3,7)           \\
		Feedforward Dimension             & 64              \\
		Transformer Cells                 & 2               \\
		Dropout                           & 0.1             \\
		\hline
	\end{tabular}
\end{table}

\section{Conclusion}
\label{sec conclusion}

In this paper, we introduced a diffusion-based generative model for counterfactual outcome prediction in longitudinal causal inference settings. By leveraging iterative denoising in a conditional diffusion process, our framework produces a full distribution over future outcomes rather than a single point estimate, addressing a key limitation of conventional time-series causal methods. Our experiments on a simulated pharmacokinetic/pharmacodynamic tumor growth dataset, where ground-truth counterfactuals are known, demonstrated that our Causal Diffusion Model (CDM) matches or improves upon state-of-the-art baselines, with particularly robust performance in distributional metrics and under high levels of time-dependent confounding.

Beyond its empirical strengths, CDM provides a principled way to capture multiple plausible future trajectories and quantify uncertainty about patient responses to potential interventions. This feature is especially critical in high-stakes domains such as healthcare, where clinicians must weigh not only the expected outcome of a treatment but also the associated risks. Moreover, our model's ability to handle high-dimensional distributions with complex temporal dependencies positions it as a promising approach for a broad class of causal time-series tasks. Furthermore, the approach, while not shown here, has an inherent ability in accounting for missing data, which is a common issue in real-world clinical datasets. Accounting for missing data is as simple as masking the missing entries in the input data, and the model will automatically learn to impute these missing values as part of the denoising process. At the same time, there is a risk that probabilistic counterfactual forecasts could be misused, leading to over-reliance on model outputs, potential biases in treatment planning, or reduced clinician oversight, so these tools should be deployed with appropriate safeguards and human‐in‐the‐loop review.

Several directions remain open for future work. First, applying this approach to large-scale real-world datasets and validating against external benchmarks or expert assessments would offer further evidence of its practical effectiveness. Second, improving the efficiency of the reverse diffusion process (e.g., via fast samplers) may enhance scalability. We hope this work sparks broader adoption of diffusion-based methods in causal inference and inspires follow-up research on generative approaches to modeling counterfactual phenomena.

\medskip

\noindent\emph{Limitations.}
Our evaluation is limited to a pharmacokinetic-pharmacodynamic tumor-growth simulator, which may not capture all real-world clinical complexities. The reverse diffusion sampling process is computationally intensive, potentially limiting real-time or large-scale applications. Scaling CDM to larger cohorts or higher-dimensional time-series may require further architectural or algorithmic optimizations. Finally, like other causal inference methods, our approach relies on sequential ignorability and positivity assumptions; violations of these in practice could bias counterfactual estimates. Due to computational constraints, we assessed seed-to-seed variability at three representative confounding levels using four random seeds, found negligible variability (standard deviations <0.01\%), and thus did not report confidence intervals for all experiments.

\end{document}